\documentclass[review]{elsarticle}
\usepackage[margin=2.5cm]{geometry}
\graphicspath{ {./figures/} }
\usepackage{hyperref}
\usepackage{float}
\usepackage{verbatim} 
\usepackage{apalike}
\restylefloat{figure}
\restylefloat{table}

\usepackage{algorithm}
\usepackage{algorithmic}
\usepackage{amsmath}
\usepackage{amsfonts}
\usepackage{multirow}
\usepackage{booktabs}
\usepackage{todonotes}

\def\bs{\boldsymbol}

\biboptions{authoryear}

\begin{document}
\begin{frontmatter}

%
%
%
%
%
%
%
%

\title{Real-time Forecasting of Time Series in Financial Markets Using Sequentially Trained Many-to-one LSTMs}

\author[label1]{Kelum Gajamannage \corref{cor1}}
\ead{kelum.gajamannage@tamucc.edu}

\author[label1]{Yonggi Park}
\ead{yonggi.park@tamucc.edu}


\cortext[cor1]{Corresponding author.}
\address[label1]{Department of Mathematics and Statistics, Texas A\&M University - Corpus Christ, 6300 Ocean Dr., Corpus Christi, TX 78412, USA \\}

\begin{abstract}
Financial markets are highly complex and volatile; thus, learning about such markets for the sake of making predictions is vital to make early alerts about crashes and subsequent recoveries. People have been using learning tools from diverse fields such as financial mathematics and machine learning in the attempt of making trustworthy predictions on such markets. However, the accuracy of such techniques had not been adequate until artificial neural network (ANN) frameworks were developed. Moreover, making accurate real-time predictions of financial time series is highly subjective to the ANN architecture in use and the procedure of training it. Long short-term memory (LSTM) is a member of the recurrent neural network family which has been widely utilized for time series predictions. Especially, we train two LSTMs with a known length, say $T$ time steps, of previous data and predict only one time step ahead. At each iteration, while one LSTM is employed to find the best number of epochs, the second LSTM is trained only for the best number of epochs to make predictions. We treat the current prediction as in the training set for the next prediction and train the same LSTM. While classic ways of training result in more error when the predictions are made further away in the test period, our approach is capable of maintaining a superior accuracy as training increases when it proceeds through the testing period. The forecasting accuracy of our approach is validated using three time series from each of the three diverse financial markets: stock, cryptocurrency, and commodity. The results are compared with those of an extended Kalman filter, an autoregressive model, and an autoregressive integrated moving average model.
\end{abstract}


\end{frontmatter}

\section{Introduction}\label{sec:intro}
Financial markets refer broadly to any marketplace that entitles the trading of securities, commodities, and other fungibles, and the financial security market includes stock market, cryptocurrency market, etc. \cite{Bahadur2019}. Among the three markets, which are stock, cryptocurrency, and commodity, the stock markets are well known to people while the other two are not. A cryptocurrency market exchanges digital or virtual currencies between peers without the need for a third party such as a bank \citep{nakamoto2008bitcoin}, but a commodity market trades raw materials such as gold and oil rather than manufactured products. These markets are both highly complex and volatile due to diverse economical, social, and political conditions \citep{qiu2020forecasting}. Learning such markets for the sake of making predictions is vital because that aids market analysts to make early alerts about crashes and subsequent recoveries so that the investors can either obtain better precautions for future crashes or gain more profit under future recoveries. Since it is unreliable and inefficient to rely only on a trader’s personal experience and intuition for the analysis and judgment of such markets, traders need smart trading recommendations derived from scientific research methods. 

The classical methods of making predictions on time series data are mostly linear statistical approaches such as linear parametric autoregressive (AR), moving average (MA), and autoregressive integrated moving average (ARIMA) \citep{zhao2018new} where they assume linear relationships between the current output and previous outputs. Thus, they often do not capture non-linear relationships in the data and cannot cope with certain complex time series. Because financial time series are nonstationary, nonlinear, and contaminated with high noise \citep{bontempi2012machine}, traditional statistical models have some limitations in predicting financial time series with high precision. Purely data-driven approaches such as Artificial Neural Networks (ANNs) are adopted to forecast nonlinear and nonstationary time series data with both high efficiency and better accuracy and have become a popular predictor due to adaptive self-learning \citep{gajamannage2022}.

Recurrent Neural Networks (RNNs) are powerful and robust types of ANNs that belong to the most promising algorithms in use because of their internal memory \citep{Yonggi2022}. This internal memory remembers its inputs and helps RNN to find solutions for a vast variety of problems \citep{ma2018comparison}. RNN is optimized with respect to its weights to fit the training data by adopting a technique called backpropagation that requires the gradient of the RNN. However, the gradient of RNN may vanish and explode during the optimization routing which hampers RNN's learning ability of long data sequences \citep{pascanu2013difficulty}. As a solution to these two problems \citep{le2016quantifying}, the LSTM architecture \citep{hochreiter1997long}, which is a special type of RNN, is often used. LSTMs are explicitly designed to learn long-term dependencies of time-dependent data  by remembering information for long periods. LSTM  performs faithful learning in applications such as speech recognition \citep{tian2017deep, kim2017residual} and text processing \citep{shih2017investigating, simistira2015recognition}. Moreover, LSTM is also suitable for complex data sequences such as stock time series extracted from financial markets because it has internal memory, has capability of customization, and is free from gradient-related issues.

We adopt a real-time iterative approach to train an LSTM that makes only one prediction for each iteration. For that, we train this LSTM with a known length, sat $T$ time steps, of previous data while setting the loss function to be the mean square error between labels and predictions. The LSTM predicts only one time step ahead during the current iteration that we treat as an observation for the next training dataset. We train the same LSTM over all the iterations where the number of iterations is equal to the number of total predictions. This real-time LSTM model is capable of incorporating every new future observation of the time series into the ongoing training process to make predictions. Since we use a sequence of observed time series to predict only one time step ahead, the prediction accuracy increases significantly. Moreover, the $T-1$ previous observations along with the current prediction predict for the next time step, and so the prediction error associated with the current prediction is further minimized as it runs through iterations. While classic ways of training result in more error when the predictions are made further away in the test period, our approach is capable of maintaining a superior accuracy as training increases when it proceeds through the testing period.

This paper is structured with four sections, namely, introduction (Sec.\ref{sec:intro}), methods (Sec.~\ref{sec:meth}), performance analysis (Sec.~\ref{sec:perf}), and discussion (Sec.~\ref{sec:disc}). In Sec.\ref{sec:intro}, first, we present the notion of real-time time series predictions. Then, we provide mathematical formulation of many-to-one LSTM architecture for sequential training. Finally, for the state-of-the-art time series prediction methods, we provide the formulation of one nonlinear statistical approach called extended Kalman filters (EKF), and two linear statistical approaches called AR and ARIMA. Sec.~\ref{sec:perf} provides a detailed analysis of the performance of our LSTM architecture against that of EKF, AR, and ARIMA using three financial stocks (Apple, Microsoft, Google), three cryptocurrencies (Bitcoin, Ethereum, Cardano), and three commodities (gold, crude oil, natural gas). We present the conclusions along with a discussion in Sec.~\ref{sec:disc}.

\begin{table}
\begin{tabular}{p{3cm}|p{11cm}}
Notation &  Description \\
\hline 
$t$ & Index for time steps\\
$T$ & Length of the training period\\
$N$ & Forecasting length\\
$L$ & Number of epochs \\
$K$ & Number of stacked LSTMs \\
$\mathcal{L}^{(t)}_l$ & Training loss at the $l$-th epoch of the $t$-th iteration\\
$\sigma$ & Sigmoid function in LSTM\\
$p$ & Order of the AR model \\
$q$ & Number of past innovations in MA model \\
$\mathcal{E}$ & Relative root mean square error \\

\hline
$\bs{x}^{(t)}$ & The observation at the $t$-th time step where $1\le t\le T$\\
$\bs{X}^{(t)}= \left[\bs{x}^{(t)}, ..., \bs{x}^{(T+t-2)}\right]$ & $t$-th input training window\\
$\hat{\bs{x}}^{(t)}$ & The prediction at the $t$-th time step where $T<t\le T+N$\\
$\bs{w}^{(t)}$ & White Gaussian noise vector with zero mean in EKF\\
$\bs{y}^{(t)}$ & Observation vector at the $t$-th time step in EKF\\
$\bs{v}^{(t)}$ & White Gaussian noise vector with zero mean in EKF\\
$\alpha_i, 1\le i\le p$ & Parameters of AR model\\
$\bs{\epsilon}^{(t)}$ & White Gaussian noise vector with zero mean in AR model\\
$\beta_i, 1\le i \le q$ & Parameters of MA model\\
$\bs{a}^{(t)}$ & $t$-th past innovation of MA model\\
$\bs{c}$ & Biased vector in ARIMA\\
$\bs{b}_i, \bs{b}_f, \bs{b}_c $ & Bias vectors in LSTM\\

\hline
$f$ & System dynamics in EKF\\
$h$ & Measurement function in EKF\\
$L^i$ & $i$-th level lag operator\\
$\Delta^D(\cdot)$ & $D$-th differential time series \\

\hline
$Q^{(t)}$, $R^{(t)}$, $P^{(t)}$ & Covariance matrices of $\bs{w}^{(t)}$, $\bs{v}^{(t)}$, $\bs{x}^{(t)}$, respectively, in EKF\\
$J_f$,  $J_h$ & Jacobian matrices of $f(\cdot)$, $h(\cdot)$, respectively, in EKF\\
$W_i, W_f, W_o$ & Weight matrices in LSTM\\
\hline
\end{tabular}
\caption{Notations used in this paper and their descriptions}
\end {table}\label{tab:nom}


\begin{table}[htp]
\begin{tabular}{p{2cm}|p{10cm}}
Abbreviations &  Description \\

\hline
LSTM & Long Short-Term Memory\\ 
KF & Kalman Filter \\
EKF & Extended Kalman Filter\\
AR & AutoRegressive\\
MA & Moving Average\\
ARMA & AutoRegressive Moving Average\\
ARIMA & AutoRegressive Integrated Moving Average\\

\end{tabular}
\caption {Abbreviations used in this paper and their descriptions}
\end{table}\label{tab:abrv}

\section{Methods}\label{sec:meth}
In this section, first, we provide technical details of the real-time time series prediction scheme. Then, we present the LSTM architecture to cater to the real-time time series prediction, and LSTM's training and predicting procedures. Moreover, we apply this real-time prediction scheme for three other time series prediction methods, namely, EKF, AR, and ARIMA. These three methods will be our state-of-the-art methods to compare the performance of LSTM. 

\subsection{Real-time time series prediction}
\begin{figure}[htp]
\centering
\includegraphics[width = .85\linewidth]{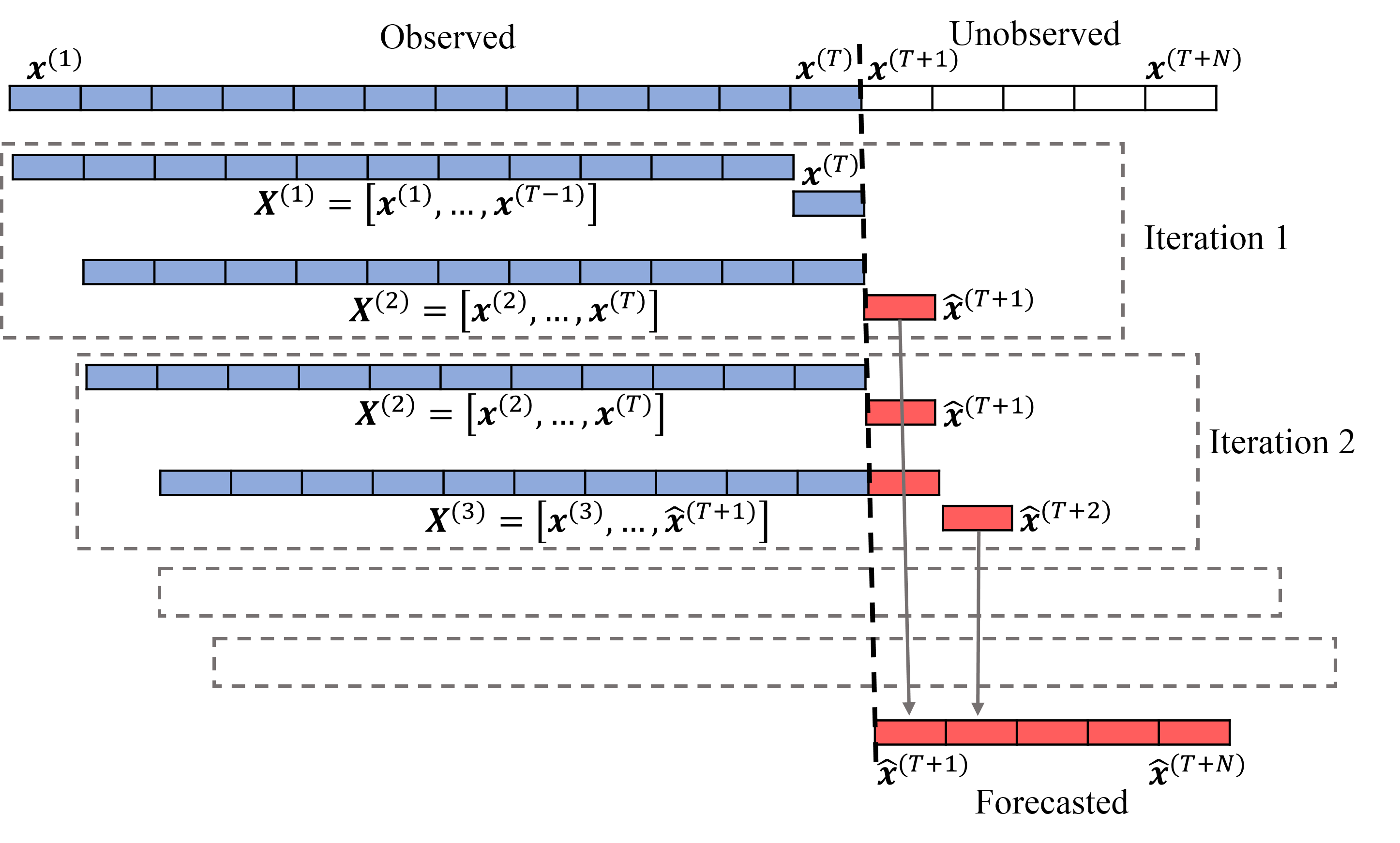}
\caption{Real-time time series prediction scheme where the currently observed time series, $\left[\bs{x}^{(1)}, \dots, \bs{x}^{(T)}\right]$ for some $T$, is shown by blue color. The unobserved future time series, $\left[\bs{x}^{(T+1)}, \dots, \bs{x}^{(T+N)}\right]$ for some $N<T$, is shown by white color while the prediction for the future is shown in red color. For the first iteration, we train the time series forecasting model with the $\bs{x}^{(T)}=\mathcal{F}\left(\bs{X}^{(1)}\right)$ where $\bs{X}^{(1)}=\left[\bs{x}^{(1)}, \dots, \bs{x}^{(T-1)}\right]$. Then, we predict for the time step $(T+1)$ as $\hat{\bs{x}}^{(T+1)}=\mathcal{F}\left(\bs{X}^{(2)}\right)$ where $\bs{X}^{(2)}=\left[\bs{x}^{(2)}, \dots, \bs{x}^{(T)}\right]$. In the second iteration, we train the same model with $\hat{\bs{x}}^{(T+1)}=\mathcal{F}\left(\bs{X}^{(2)}\right)$ where $\bs{X}^{(2)}=\left[\bs{x}^{(2)}, \dots, \bs{x}^{(T)}\right]$ and predict for the time step $(T+2)$ as $\hat{\bs{x}}^{(T+2)}=\mathcal{F}\left(\bs{X}^{(3)}\right)$ where $\bs{X}^{(3)}=\left[\bs{x}^{(3)}, \dots, \hat{\bs{x}}^{(T+1)}\right]$. We keep on doing this process until all the predictions are performed.}
 \label{fig:real_time}
\end{figure}

We adopt a ``sequential” approach to efficiently train time series models and predict for future. For a fixed-length input sequential data, the model is set to predict only one future time step at an iteration where the process runs until the required length of the prediction is performed. This real-time prediction approach is capable of incorporating every new data point of the time series into the ongoing training process to make predictions for the next time step. Let, the current observed time series is $[\bs{x}^{(1)}, \dots, \bs{x}^{(T)}]$ for some $T$, the unobserved future portion of the time series is $\left[\bs{x}^{(T+1)}, \dots, \bs{x}^{(T+N)}\right]$ for some $N<T$, and the time series model is $\mathcal{F}$, see Fig.~\ref{fig:real_time}. For the first iteration, we train the time series forecasting model with the $x^{(T)}=\mathcal{F}\left(\bs{X}^{(1)}\right)$ where $\bs{X}^{(1)}=\left[\bs{x}^{(1)}, \dots, \bs{x}^{(T-1)}\right]$. Then, we predict for the time step $(T+1)$, denoted by $\hat{\bs{x}}^{(T+1)}$, as $\hat{\bs{x}}^{(T+1)}=\mathcal{F}\left(\bs{X}^{(2)}\right)$ where $\bs{X}^{(2)}=\left[\bs{x}^{(2)}, \dots, \bs{x}^{(T)}\right]$. In the second iteration, we train the same model $\mathcal{F}$ with $\hat{\bs{x}}^{(T+1)}=\mathcal{F}\left(\bs{X}^{(2)}\right)$ where $\bs{X}^{(2)}=\left[\bs{x}^{(2)}, \dots, \bs{x}^{(T)}\right]$ and predict for the time step $(T+2)$, denoted by $\hat{\bs{x}}^{(T+2)}$,  as $\hat{\bs{x}}^{(T+2)}=\mathcal{F}\left(\bs{X}^{(3)}\right)$ where $\bs{X}^{(3)}=\left[\bs{x}^{(3)}, \dots, \hat{\bs{x}}^{(T+1)}\right]$. We keep on doing this process until the predictions are performed for all the $N$ time steps .

\subsection{Many-to-one LSTM architecture with sequential training}
\begin{figure}[htp]
\includegraphics[width = .85\linewidth]{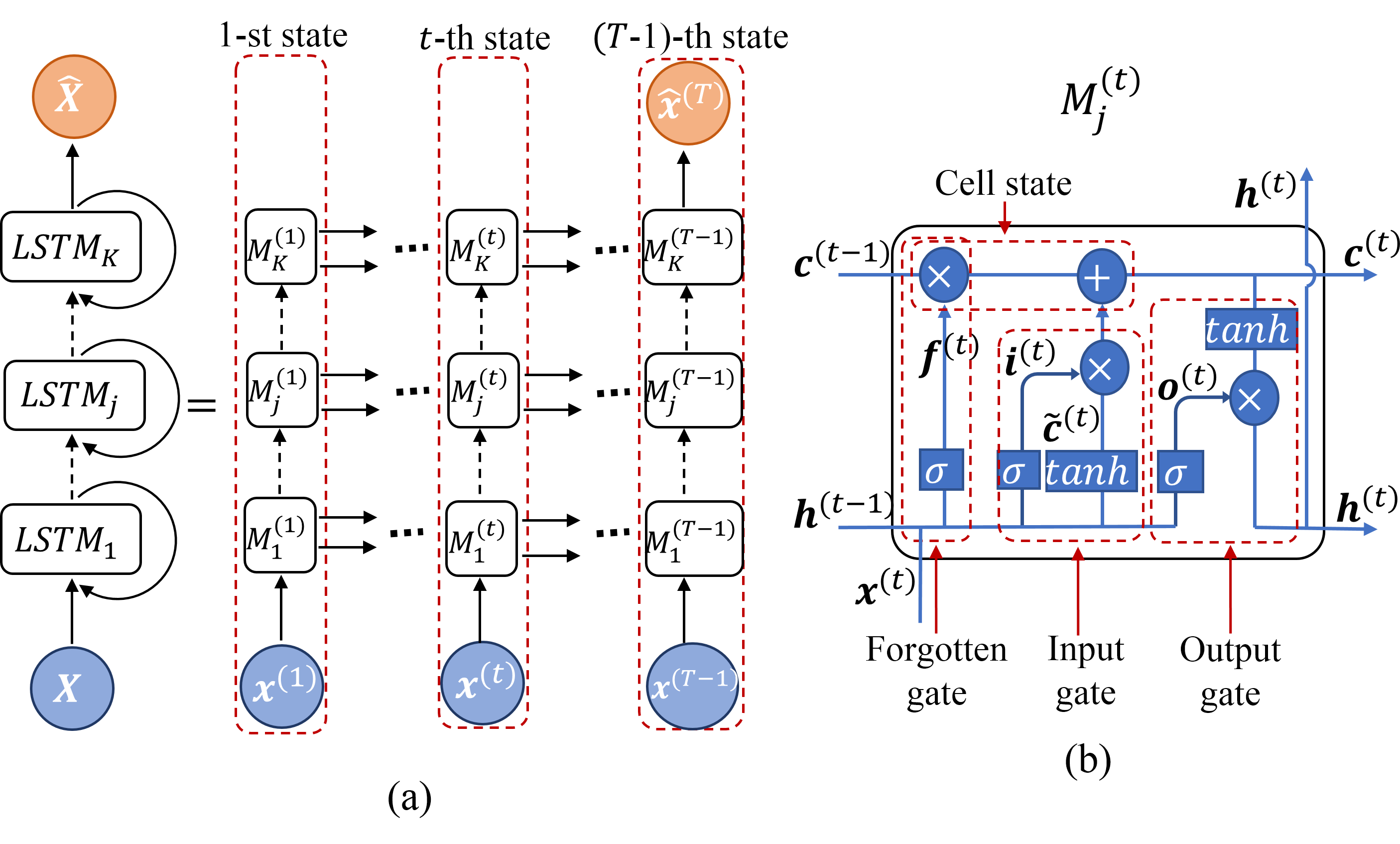}
\centering
\caption{$K$-stacked LSTMs for many-to-one forecasting of a single feature time series where each LSTM is a collection of recurrent modules, denoted as $M$'s. (a) the left figure shows a folded version of the artificial neural network (ANN) whereas the right figure shows its unfolded version of it. Here, the input for the ANN is $\bs{X}=\left[\bs{x}^{(1)}, \dots, \bs{x}^{(t)}, \dots, \bs{x}^{(T-1)}\right]$ and output from that is $\hat{\bs{x}}^{(T)}$. (b) Each $M$ filters information through four hidden layers where three of them are gates, namely, forgotten, input, and output, and the other is called the cell state. The forgotten gate resets the content of the memory cell, the input gate decides what new information is stored in the memory cell, the cell state stores long-term information in the memory, and the output gate sends out a filtered version of the memory cell's stored information from the $M$. Operations in an $M$ are given as follows: $\bs{f}^{(t)} = \sigma\left(W_f \cdot [\bs{h}^{(t-1)}, \bs{x}^{(t)}] + \bs{b}_f\right)$, $\bs{i}^{(t)} = \sigma\left(W_i \cdot [\bs{h}^{(t-1)}, \bs{x}^{(t)}] + \bs{b}_i\right)$,  $\tilde{\mathbf{c}}^{(t)} = tanh\left(W_C \cdot [\bs{h}^{(t-1)}, \bs{x}^{(t)}] + \bs{b}_c\right)$, $\bs{c}^{(t)}=\bs{f}^{(t)}\odot \bs{c}^{(t-1)}\oplus\bs{i}^{(t)}\odot \tilde{\bs{c}}^{(t)}$, $\bs{o}^{(t)}=\sigma\left(W_o \cdot [\bs{h}^{(t-1)}, \bs{x}^{(t)}] + \bs{b}_o\right)$, and $\bs{h}^{(t)} = \bs{o}^{(t)} \odot tanh\left(\bs{c}^{(t)}\right)$, where $\odot$ and $\oplus$ are point-wise multiplication and point-wise addition, respectively.}
\label{fig:lstmGate}
\end{figure}

Since we make predictions only for one time step ahead at a time for an input time series, the LSTM architecture implemented here is the many-to-one type, see Fig.~\ref{fig:lstmGate}(a) for $K$-stacked LSTM architecture. An LSTM consists of a series of nonlinear recurrent modules, denoted as $M^{(t)}_j$ for $t=1,\dots,T$ and $j=1,\dots,N$ in Fig.~\ref{fig:lstmGate}, where each module processes data related to one time step. LSTM introduces a memory cell, a special type of the hidden state, that has the same shape as the hidden state which is engineered to record additional information. Each recurrent module in an LSTM filters information through four hidden layers where three of them are gates, namely, forgotten gate, input gate, and output gate, and the other is called the cell state that maintains and updates long-term memory, see \ref{fig:lstmGate}(b). 

The forgotten gate resets the content of the memory cell by deciding what information should be forgotten or retained. This gate produces a value between zero and one where zero means completely forgetting the previous hidden state and one means completely retaining that. Information from the previous hidden state, i.e., $\bs{h}^{(t-1)}$, and the information from the current input, i.e., $\bs{x}^{(t)}$, is passed through the $sigmoid$ function, denoted as $\sigma$, according to
\begin{equation}\label{eqn:forg}
\bs{f}^{(t)} = \sigma\left(W_f \cdot [\bs{h}^{(t-1)}, \bs{x}^{(t)}] + \bs{b}_f\right),
\end{equation}
where $W_f$ and $\bs{b}_f$ are weighting matrix and biased vector, respectively. The input gate consisting of two components decides what new information is to be stored in the cell state. The first component is a $sigmoid$ layer that decides which values to be updated based on the previous hidden state and the information from the current input such that
\begin{equation}\label{eqn:inpt}
\bs{i}^{(t)} = \sigma\left(W_i \cdot [\bs{h}^{(t-1)}, \bs{x}^{(t)}] + \bs{b}_i\right),
\end{equation} 
where $W_i$ and $\bs{b}_i$ are weighting matrix and biased vector, respectively. The next component is a $tanh$ layer that creates a vector of new candidate values, $\tilde{c}^{(t)}$, based on the previous hidden state and the information from the current input as
\begin{equation}\label{eqn:cell1}
\tilde{\mathbf{c}}^{(t)} = tanh\left(W_c \cdot [\bs{h}^{(t-1)}, \bs{x}^{(t)}] + \bs{b}_c\right),
\end{equation}
where $W_c$ and $\bs{b}_c$ are weighting matrix and biased vector, respectively. 

Cell state updates the LSTM's memory with new long-term information. For that, first, it multiplies point wisely the old cell state $c^{(t-1)}$ by the forgetting state $\bs{f}^{(t)}$, i.e., $\bs{f}^{(t)}\odot \bs{c}^{(t-1)}$, to assure that the information retains from the old cell state is what is allowed by the forgetting state. Then, we add the pointwise product $\bs{i}\odot\tilde{c}^{(t)}$ into $\bs{f}^{(t)}\odot \bs{c}^{(t-1)}$, i.e., 
\begin{equation}\label{eqn:cell2}
\bs{c}^{(t)}=\bs{f}^{(t)}\odot \bs{c}^{(t-1)}\oplus\bs{i}^{(t)}\odot \tilde{\bs{c}}^{(t)},
\end{equation}
as the information from the current input state which is found relevant by the ANN. The output gate determines the value of the next hidden state with the information from the current cell state, current input state, and previous hidden state. First, a $sigmoid$ layer decides how much of the current input and the previous hidden state are going to output. Then, the current cell state is passed through the $tanh$ layer to scale the cell state value between -1 and 1. Thus, the output $\bs{h}^{(t)}$ is
\begin{equation}\label{eqn:outp}
\bs{h}^{(t)} =\bs{o}^{(t)}\odot tanh\left(\bs{c}^{(t)}\right), \ \text{with} \ \   
\bs{o}^{(t)}=\sigma\left(W_o \cdot [\bs{h}^{(t-1)}, \bs{x}^{(t)}] + \bs{b}_o\right),
\end{equation}
where $W_o$ and $\bs{b}_o$ are weighting matrix and biased vector, respectively. Based upon $\bs{h}^{(t)}$, the network decides which information from the current hidden state should be carried out to the next hidden state where the next hidden state is used for prediction. To conclude, the forget gate determines which relevant information from the prior steps is needed. The input gate decides what relevant information can be added from the current cell state, and the output gates finalize the input to the next hidden state.

\subsubsection{Optimization of LSTM}
Training an LSTM is the process of minimizing a relevant reconstruction error function, also called loss function, with respect to weights and bias vectors of Eqns.~\eqref{eqn:forg}, \eqref{eqn:inpt}, \eqref{eqn:cell1}, \eqref{eqn:cell2}, and \eqref{eqn:outp}. Such a minimization problem is implemented in four steps: first, forward propagation of input data through the ANN to get the output; second, calculate the loss, between forecasted output and the true output; third, calculate the derivatives of the loss function with respect to the LSTM's weights and bias vectors using backpropagation through time (BTT) \cite{werbos1990backpropagation}; and fourth, adjusting the weights and bias vectors by gradient descent method \cite{gruslys2016memory}.

BTT unrolls backward all the dependencies of the output onto the weights of the ANN \cite{manneschi2020alternative}, which is represented from left side to right side in Fig.~\ref{fig:lstmGate}(a). At each iteration, say $t\in[1,N+1]$, we train the LSTM by only one instance of  input-label where the input is $\bs{X}^{(t)} =\left[\bs{x}^{(t)},\dots,\bs{x}^{(T)},\hat{\bs{x}}^{(T+1)},\dots,\hat{\bs{x}}^{(T+t-2)}\right]$ and the label is $\hat{\bs{x}}^{(T+t-1)}$. Due to this process, at the $t$-th iteration, the ANN is trained with $t$-th input-label instance and predicts for the ($T+t-1$)-th time step. Thus, we formulate the loss function at the $t$-th iteration of the LSTM as the relative mean square error, 
\begin{equation} \label{eq:mse}
\mathcal{L}^{(t)} = \frac{\left\|\hat{\bs{x}}^{(T+t-1)} - \bs{x}^{(T+t-1)}\right\|_F^2}{\left\|\bs{x}^{(T+t-1)}\right\|_F^2},
\end{equation}
where $F$ denotes the Frobenius norm and $\hat{\bs{x}}^{(T+t-1)}$ is the output of the LSTM for the input $\bs{X}^{(t-1)}$. We use BTT to compute the derivatives of Eqn.~\eqref{eq:mse} with respect to the weights and bias vectors. We update the weights using the gradient descent-based method, called Adaptive Moment Estimation (ADAM) \cite{kingma2014adam}. ADAM is an iterative optimization algorithm used in recent machine learning algorithms to minimize loss functions where it employs the averages of both the first-moment gradients and the second-moment of the gradients for computations. It generally converges faster than standard gradient descent methods and saves memory by not accumulating the intermediate weights. 

\begin{figure}[htp]
\centering
\includegraphics[width = 3.5in]{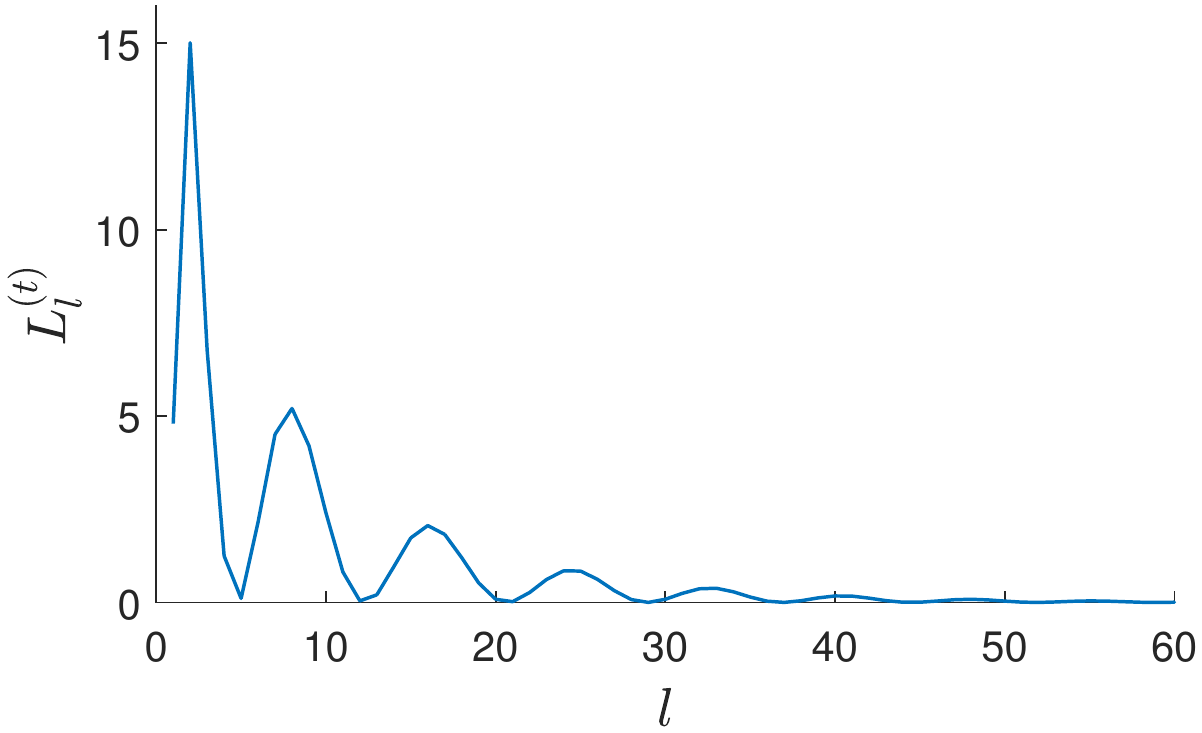}
\caption{Non-convex behavior of the loss function. We apply our training scheme to train an LSTM with closing prices of the Apple stock. We input a sequence of 1227 days of prices into the LSTM and generate the price for the 1228-th day. The loss is computed as the relative mean square error between the predicted price and the observed prices for the 1228-th day. We proceed with this single-day training 60 times, also called epochs, where the loss for the $l$-th epoch is denoted as $L^{(t)}_l$.}
\label{fig:semicon}
\end{figure}

To assure better convergence of the loss function, we integrate epochs into the training process in a unique way that we explain here for the $t$-th iteration. However, if the loss function is non-convex or semi-convergence choosing the best number of epochs is challenging. Fig.~\ref{fig:semicon} illustrates the non-convex behavior of an LSTM's loss function that is trained with the closing prices of the Apple stock. Here, we input a sequence of 1227 days of prices into the LSTM and generate the price for the 1228-th day where  the loss is computed as the relative mean square error between the predicted price and the observed prices for the 1228-th day. We proceed with this single-day training for 60 epochs as shown in Fig.~\ref{fig:semicon}. Since the loss varies non-convexly with respect to epochs, we came up with a unique way of training the LSTM through epochs. Particularly, we maintain two LSTMs, denoted as $LSTM_1$ and $LSTM_2$, that are trained through each iteration. We assume that those two LSTMs corresponding to the $(t-1)$-th iteration are given for the $t$-th iteration. For the $t$-th iteration, we train $LSTM_1$ with the input $\bs{X}^{(t)}$ and the label $\bs{x}^{T+t-1}$ for fix number of epochs, say $L$. Here, we record $LSTM_1$'s optimum weights and biased vectors corresponding to each of the epoch. We reformulate $LSTM_2$ with the weights and biased vectors corresponding to the least loss among $L$ epochs. Finally, we redefine $LSTM_1$ as $LSTM_2$ and proceed to the $(t+2)$-th iteration. Algorithm~\ref{algo:lstm} summarizes the training and prediction procedure of our sequentially trained many-to-one LSTM scheme.

\begin{algorithm}[htp]
\caption{: Many-to-one LSTM architecture with sequential training.\newline 
Denotation: $\bs{X}^{(1)} =\left[\bs{x}^{(1)},\dots,\bs{x}^{(T-1)}\right]$, $\bs{X}^{(2)} =\left[\bs{x}^{(2)},\dots,\bs{x}^{(T)}\right]$, and 
$\bs{X}^{(t)} =\left[\bs{x}^{(t)},\dots,\bs{x}^{(T)},\hat{\bs{x}}^{(T+1)},\dots,\hat{\bs{x}}^{(T+t-2)}\right]$ for $3\le t\le N+1$.  \newline
Input: training time series $\left(\left[\bs{x}^{(1)}, \dots, \bs{x}^{(T)}\right]\right)$; forecast length ($N$); number of maximum epochs ($L$).\newline 
Output:  time series forecast $\left(\left[\hat{\bs{x}}^{(T+1)}, \dots, \hat{\bs{x}}^{(T+N)}\right]\right)$; trained $LSTM$ ($LSTM_1$ or $LSTM_2$).\newline}
\begin{algorithmic}[1]
\STATE Initialization: two $LSTM$s, denoted as $LSTM_1$ and $LSTM_2$, with the weights $W_f=W_i=W_c=W_o=0$ and biased vectors $\bs{b}_f=\bs{b}_i=\bs{b}_c=\bs{b}_o=\bs{h}^{(0)}=\textbf{0}$.\hspace{-1cm}
\FOR{$t\in [1,N+1]$} 
\FOR{$l\in[1,L]$} 
\STATE Compute $\hat{\bs{x}}^{(T+t-1)} = LSTM_1\left(\bs{X}^{(t)}\right)$ according to the map in Fig~\ref{fig:lstmGate}(a).
\STATE Minimize $L^{(t)}_l = \left\|\hat{\bs{x}}^{(T+t-1)} - \bs{x}^{(T+t-1)}\right\|_F^2\Big/\left\|\bs{x}^{(T+t-1)}\right\|_F^2$ using BTT with respect to the weights $W_i$, $W_i$, $W_c$, and $W_o$, and bias vectors $\bs{b}_f$, $\bs{b}_i$, $\bs{b}_c$, and $\bs{b}_o$ of the composite representation of the functions in Eqns.~\eqref{eqn:forg}, \eqref{eqn:inpt}, \eqref{eqn:cell1}, \eqref{eqn:cell2}, and \eqref{eqn:outp}.
\STATE Record $L^{(t)}_l$ along with $W_i$, $W_i$, $W_c$, $W_o$, $\bs{b}_f$, $\bs{b}_i$, $\bs{b}_c$, and $\bs{b}_o$.
\STATE Update the weights and bias vectors of $LSTM_1$ using the gradient descent-based method ADAM.
\ENDFOR
\STATE Reformulate $LSTM_2$ with $W_i$, $W_i$, $W_c$, $W_o$, $\bs{b}_f$, $\bs{b}_i$, $\bs{b}_c$, and $\bs{b}_o$ corresponding to the least $\left\{L^{(t)}_l\Big\vert \ \forall l\right\}$.
\STATE Replicate $LSTM_2$ and define it as $LSTM_1$.
\STATE Forecast for $(T+t)$-th time step, i.e., $\hat{\bs{x}}^{(T+t)} = LSTM_1\left(\bs{X}^{(t+1)}\right)$
\ENDFOR
\end{algorithmic}
\label{algo:lstm}
\end{algorithm}

\subsection{State-of-the-art methods}
Here, we present three state-of-the-art time series prediction methods, namely, extended Kalman filter (EKF), autoregression, and autoregressive integrated moving average (ARIMA), that we use to validate the performance of our LSTM scheme. Here, we utilize the same sequential training as we did for LSTMs to make real-time predictions on the same financial time series. 

\subsubsection{Extended Kalman Filter (EKF)}
EKF is a nonlinear version of the standard Kalman filter where the formulation of EKF is based on the linearization of both the state and the observation equations. In an EKF, the state Jacobian and the measurement Jacobian replace the state transition matrix and the measurement matrix,  respectively, of a linear KF (\cite{valade2017study}). This process essentially linearizes the non-linear function around the current estimate.  Linearization enables the propagation of both the state and state covariance in an approximately linear format. Here, the extended Kalman filter is presented in three steps, namely, dynamic process, model forecast step, and data assimilation step. 

\paragraph{Dynamic Process} 
Here, we present both the state model and the observation model of a nonlinear dynamic process. The current state, $\bs{x}^{(t+1)}$, is modeled as a sum of the nonlinear function of the previous state, $\bs{x}^{(t)}$, and the noise, $\bs{w}^{(t)}$, as
\begin{equation}\label{eq2}
\bs{x}^{(t+1)} = f\left(\bs{x}^{(t)}\right) + \bs{w}^{(t)},
\end{equation}
where $\bs{x}^{(t)}$, $\bs{w}^{(t)} \in \mathbb{R}^n$. Here, the random process $\{\bs{w}^{(t)}\}$ is Gaussian white noise with zero mean and covariance matrix of $Q^{(t)}=E\left[\bs{w}^{(t)} \left(\bs{w}^{(t)}\right)^T\right]$. The initial state $\bs{x}^{(0)}$ is a random vector with known mean $\bs{\mu}_0=E\left(\bs{x}^{(0)}\right)$ and covariance $P^{(0)}=E\left[(\bs{x}^{(0)}-\bs{\mu}_0)(\bs{x}^{(0)}-\bs{\mu}_0)^T\right]$. The Jacobian of the predicted state with respect to the previous state, denoted as $J_f$, is obtained by partial derivatives as $J_f = f_{\bs{x}}(\cdot)$.

The current observation, $\bs{y}^{(t+1)}$, is modeled as a sum of the nonlinear function of the current state, $\bs{x}^{(t+1)}$, and the noise, $\bs{v}^{(t+1)}$, as
\begin{equation}\label{eq9}
\bs{y}^{(t)} = h\left(\bs{x}^{(t)}\right) + \bs{v}^{(t)},  
\end{equation}
where $\bs{y}^{(t+1)}$, $\bs{v}^{(t+1)} \in \mathbb{R}^n$. Here, the random process $\{\bs{v}^{(t+1)}\}$ is Gaussian white noise with zero mean and covariance matrix of $R^{(t)}=E\left[\bs{v}^{(t)} \left(\bs{v}^{(t)}\right)^T\right]$. The Jacobian of the predicted observation with respect to the previous state, denoted as $J_h$, is obtained by partial derivatives as $J_h = h_{\bs{x}}(\cdot)$.

\paragraph{Model Forecast Step}
The state Jacobian and the measurement Jacobian replace linear KF's state transition matrix and the measurement matrix, respectively \cite{valade2017study}. Let, the initial estimates of the state and the covariance are $\bs{x}^{(0|0)}$ and $P^{(0|0)}$, respectively. The state and the covariance matrix are propagated to the next step using 
\begin{eqnarray}\label{eq16}
\hat{\bs{x}}^{(t+1)} \approx f\left(\hat{\bs{x}}^{(t)}\right)
\end{eqnarray}
and 
\begin{eqnarray}
\label{eq17}
P^{(t+1)} = J_f\left(\hat{\bs{x}}^{(t)}\right) P^{(t)} \left[J_f\left(\hat{\bs{x}}^{(t)}\right)\right]^T + Q^{(t)},
\end{eqnarray}
respectively. 

\paragraph{Data Assimilation Step} 
The measurement at the $t+1$ step is given by 
\begin{eqnarray}\label{eq18}
\bs{y}^{(t+1)} \approx  h\left(\hat{\bs{x}}^{(t+1)}\right)
\end{eqnarray}
Use the difference between the actual measurement and the predicted measurement to correct the state at the $t+1$ step. To correct the state, the filter must compute the Kalman gain. First, the filter computes the measurement prediction covariance (innovation) as
\begin{eqnarray}\label{eq19}
S^{(t+1)} = J_h\left(\bs{x}^{(t+1)}\right) P^{(t+1)} \left[J_h\left(\bs{x}^{(t+1)}\right)\right] + R^{(t+1)}
\end{eqnarray}
Then, the filter computes the Kalman gain as
\begin{eqnarray}\label{eq20}
K_g^{(t+1)} = P^{(t+1)} \left[J_h\left(\bs{x}^{(t+1)}\right)\right]^T \left[S^{(k+1)}\right]^{-1} 
\end{eqnarray}
The filter corrects the predicted estimate by using observation. The estimate, after the correction using the observation $\bs{y}^{(t+1)}$, is
\begin{eqnarray}
\label{eq21}
\begin{split}
\hat{\bs{x}}^{(t+1)} = \bs{x}^{(t+1)} + K^{(t+1)} \left[\bs{y}^{(t+1)} -h\left( \bs{x}^{(t+1)}\right)\right]\\
P^{(t+1)} = \left[I - K_g^{(t+1)} J_h\left(\hat{\bs{x}}^{(t+1)}\right)\right] P^{(t+1)}
\end{split}
\end{eqnarray}
The corrected state is often called the a posteriori estimate of the state, because it is derived after including the observation.

\subsubsection{Autoregression (AR) model}
Many observed time series exhibit serial autocorrelation which is known to be the linear association between lagged observations. The AR model predicts the value for the current time step, $\bs{x}^{(t)}$, based on a linear relationship between $p$-recent observations, $\bs{x}^{(t-1)}$, $\bs{x}^{(t-2)}$, $\dots$, $\bs{x}^{(t-p)}$, where this $p$ is known as the order of the model \cite{Geurts1977}. Let, $\alpha_1, \dots, \alpha_p\in\mathbb{R}$ are the coefficients, order $p$ AR model is given by 
\begin{eqnarray}\label{eq:ar1}
\bs{x}^{(t)} = \bs{c} + \alpha_1 \bs{x}^{(t-1)} + \alpha_2 \bs{x}^{(t-2)} + ... + \alpha_p \bs{x}^{(t-p)} + \bs{\epsilon}^{(t)},
\end{eqnarray}
where $\bs{\epsilon}^{(t)}$ is uncorrelated noise with a zero mean. Let, the lag operator polynomial notation is $L^i\bs{x}^{(t)}=\bs{x}^{(t-i)}$. We define order $p$ autoregression lag operator polynomial as $\alpha(L)=(1-\alpha_1 L - \dots -\alpha_p L^p)$. Thus, AE model is given by
\begin{eqnarray}
\alpha(L)\bs{x}^{(t)} = c + \bs{\epsilon}^{(t)}.
\end{eqnarray}
The solution for the AR model is given by
\begin{eqnarray}
\bs{x}^{(t)} = \alpha^{-1}(L)\left(c + \bs{\epsilon}^{(t)}\right).
\end{eqnarray}

\subsubsection{Autoregressive integrated moving average (ARIMA) model}
ARIMA model is made by combining a differential version of AR model into a moving average (MA) model. MA model captures serial autocorrelation in a time series $\bs{x}^{(1)},\dots,\bs{x}^{(t)},\dots,\bs{x}^{(T)}$ by expressing the conditional mean of $\bs{x}^{(t)}$ as a function of past innovations, $\bs{a}^{t()}, \bs{a}^{(t-1)}, ... , \bs{a}^{(t-q)}$. An MA model that depends on $q$ past innovations is called an MA model of order $q$, denoted by MA($q$).
In general, the MA($q$) model can be represented by the formula
\begin{eqnarray}
\bs{x}^{(t)} = \bs{\mu} +  \bs{a}^{(t)} + \beta_1 \bs{a}^{(t-1)} + ... + \beta_q \bs{a}^{(t-q)},
\end{eqnarray}
where $\bs{a}^{(t)}$'s are uncorrelated innovation processes with a zero mean and $\mu$ is the unconditional mean of $\bs{x}^{(t)}$ for all $t$.

For some observed time series, a higher order AR or MA model is needed to capture the underlying process well. In this case, a combined ARMA model can sometimes be a parsimonious choice. An ARMA model expresses the conditional mean of $\bs{x}^{(t)}$ as a function of both recent observations, $\bs{x}^{(t-1)}, \bs{x}^{(t-2)}, \dots, \bs{x}^{(t-p)}$, and recent innovations, 
$\bs{a}^{(t)}, \bs{a}^{(t-1)}, \dots , \bs{a}^{(t-q)}$. The ARMA model with AR degree of $p$ and MA degree of $q$ is denoted by ARMA ($p, q$), which is given by 
\begin{eqnarray}\label{eq:arma}
\begin{split}
\bs{x}^{(t)} = \bs{c} + \alpha_1 \bs{x}^{(t-1)} + \alpha_2 \bs{x}^{(t-2)} + \dots + \alpha_p \bs{x}^{(t-p)} \\
+ \beta_0 \bs{a}^{(t)} + \beta_1 \bs{a}^{(t-1)} + \dots + \beta_q \bs{a}^{(t-q)},
\end{split}
\end{eqnarray}
\cite{arima}. 

The ARIMA process generates nonstationary series that are integrated of order $D$ where that nonstationary process can be made stationary by taking $D$ differences. A series that can be modeled as a stationary ARMA($p,q$) process after being differenced $D$ times is denoted by ARIMA($p,D,q$), which is given by 
\begin{equation}
\begin{split}
\Delta^D \bs{x}^{(t)} = \mu + \alpha_1 \Delta^D \bs{x}^{(t-1)} + \alpha_2 \Delta^D \bs{x}^{(t-2)} + ... + \alpha_p \Delta^D \bs{x}^{(t-p)} \\ + \bs{a}^{(t)} - \beta_1 \bs{a}^{(t-1)} - \beta_2 \bs{a}^{(t-2)} - ... - \beta_q \bs{a}^{(t-q)}, 
\end{split}
\end{equation}
where $\Delta^D \bs{x}^{(t)}$ denotes a $D$-th differential time series, $\bs{a}^{(t)}$'s are uncorrelated innovation processes with a zero mean, and $\mu$ is the unconditional mean of $\bs{x}^{(t)}$ for all $t$ \citep{newbold1983arima}. With the lag operator $L^i \bs{x}^{(t)} = \bs{x}^{(t-i)}$, the ARIMA model can be written as 
\begin{eqnarray}
\alpha (L) {(1-L)}^D \bs{x}^{(t)} = c + \beta(L) \bs{a}^{(t)}
\end{eqnarray}
where $\alpha(L) = (1 - \alpha_1 L - \dots -\alpha_p L^p)$ and $\beta(L) = (1 + \beta_1 L + \dots +\beta_q L^q)$. Thus, the solution for ARIMA model is given by
\begin{eqnarray}
\bs{x}^{(t)} = \left(\alpha (L) {(1-L)}^D\right)^{-1} \left(c + \beta(L) \bs{a}^{(t)}\right).
\end{eqnarray}

\section{Performance Analysis} \label{sec:perf}
The performance analysis of LSTM is conducted using nine financial time series obtained from three markets, namely, stocks, cryptocurrencies, and commodities. We chose Apple, Google, and Microsoft for stocks;  Bitcoin, Ethereum, Cardano for cryptocurrencies; and Gold, Oil, and Natural Gas for commodities. These diverse examples validate the broad applicability of LSTMs in analyzing and predicting financial time series.

We follow the procedure given in Fig.~\ref{fig:real_time} to train the real-time many-to-one LSTM architecture given in Fig.~\ref{fig:lstmGate}. Setting the LSTM to run for a specific number of epochs and then using that trained network to make predictions often do not perform the best training and then do not perform accurate predictions since the loss function undergoes semi-convergence as shown in Fig.~\ref{fig:semicon}. To avoid this issue; first, we train the LSTM for 100 epochs; second, we compute the best number of epochs associated with the least loss; and finally, train again a new LSTM with that many epochs. Moreover, the parameter choices for the training length and prediction length are shown in Table~\ref{tab:para}.
\begin{center}
\begin{table}[htp]
\begin{center}
\caption{Parameter choices for the training length, prediction length, and number of epochs used in real-time many-to-one LSTMs.}
\label{tab:para}
\begin{tabular}{|p{2.0cm}|c|c|c|}
\toprule
time series  & Training length  & Prediction length \\
& ($T$) & ($N$) \\ \midrule
Apple       &1228&30 \\
Microsoft           & 1228    & 30 \\ 
Google        &  1228    & 30 \\ 
\midrule
Bitcoin      &1064      & 30 \\
Ethereum       & 1064  & 30 \\
Cardano          & 1064 &30 \\
\midrule
Oil           & 8248& 200 \\
Natural gas & 5802   &150 \\
Gold            &816&  30 \\
\bottomrule
\end{tabular}
\end{center}
\end{table}
\end{center}
Now, we incorporate the same one-day recursive prediction procedure in Fig.~\ref{fig:real_time} into the other three state-of-the-art methods, namely, EKF, AR, and ARIMA, to predict the above financial time series. After a trial and error process, we found that the best $p$'s of AR are 300, 400, and 400, for Apple, Microsoft, and Google, respectively; and the best ($p, D, q$)'s of ARIMA are (10, 0, 2), (10, 2, 1), and (0, 1 , 1), for Apple, Microsoft, and Google, respectively. Then, the best $p$'s of AR were found to be 100, 100, and 300, for Bitcoin, Ethereum, and Cardano, respectively; and the best ($p, D, q$)'s of ARIMA were found to be (6, 0, 2), (6, 1, 1), and (8, 2 , 1), for Bitcoin, Ethereum, and Cardano, respectively. Finally, the best $p$'s of AR were 200, 200, and 100, for Oil, Natural gas, and Gold, respectively; and the best ($p, D, q$)'s of ARIMA were (4, 1, 1), (10, 1, 2), and (8, 2 , 0), for Oil, Natural gas, and Gold, respectively. Thus, we set the methods with the best parameter values and executed them with the corresponding time series.

We compute the mean of the relative absolute difference between the predicted and the observed time series for the prediction period using
\begin{equation}
\mathcal{E} = \frac{1}{N}\sum^{T+N}_{t = T+1}\frac{\big\|\hat{\bs{x}}^{(t)}-\bs{x}^{(t)}\big\|_2}{\big\|\bs{x}^{(t)}\big\|_2},
\end{equation} as an error measure of the prediction that we show in Table~\ref{tab:rmse}. Hereby, we observe that the order of the best to the worst prediction performance is LSTM, EKF, AR, and ARIMA.

\begin{table}[htp]
\begin{center}
\caption{This table shows the prediction error, quantified as the means of the relative difference between the predicted and the observed time series for the prediction period, of four methods LSTM, EKF, AR, and ARIMA. The analysis is conducted on three stocks Apple, Microsoft, and Google; three cryptocurrencies, Bitcoin, Ethereum, and Cardano; and three commodities, oil, natural gas, and gold. On average, LSTM performs 17 times better than EKF, 7 times better than AR, and 4 times better than ARIMA. Moreover, the average prediction errors of LSTM are 0.05, 0.22, and 0.14 for stocks, cryptocurrencies, and commodities, respectively.}
\label{tab:rmse}
\begin{tabular}{|p{2.0cm}|c|c|c|c|}
\toprule
time series   & LSTM &           EKF           & AR & ARIMA     \\ \midrule
Apple       &  $5.9*10^{-2}$         & $4.5*10^{-1}$       & $2.5*10^{-1}$ &  $7.5*10^{-2}$\\
Microsoft   &$5.5*10^{-2}$  & $4.2*10^{-1}$  & $4.3*10^{-1}$  & $6.3*10^{-2}$  \\ 
Google        &  $3.5*10^{-2}$   & $4.7*10^{-1}$       & $4.7*10^{-1}$ & $5.5*10^{-2}$      \\ 
\midrule
Bitcoin      &    $2.1*10^{-1}$     &  $2.7*10^{0}$  & $4.0*10^{-1}$ & $3.8*10^{-1}$     \\
Ethereum     &  $1.8*10^{-1}$   & $2.9*10^{0}$    & $8.1*10^{-1}$ & $1.0*10^{0}$      \\
Cardano          &  $2.7*10^{-1}$  & $4.3*10^{0}$  & $1.2*10^{0}$  & $1.7*10^{0}$  \\
\midrule
Oil           &  $1.5*10^{-1}$  & $4.6*10^{0}$   & $6.3*10^{-1}$ & $3.0*10^{-1}$        \\
Natural gas &  $2.0*10^{-1}$ & $3.4*10^{0}$   & $8.0*10^{-1}$  & $1.0*10^{0}$       \\
Gold            &    $7.7*10^{-2}$ & $2.4*10^{0}$     & $1.2*10^{0}$  & $1.1*10^{0}$    \\
\bottomrule
\end{tabular}
\end{center}
\end{table}

\begin{figure}[htp]
\centering
\includegraphics[width = 6.5in]{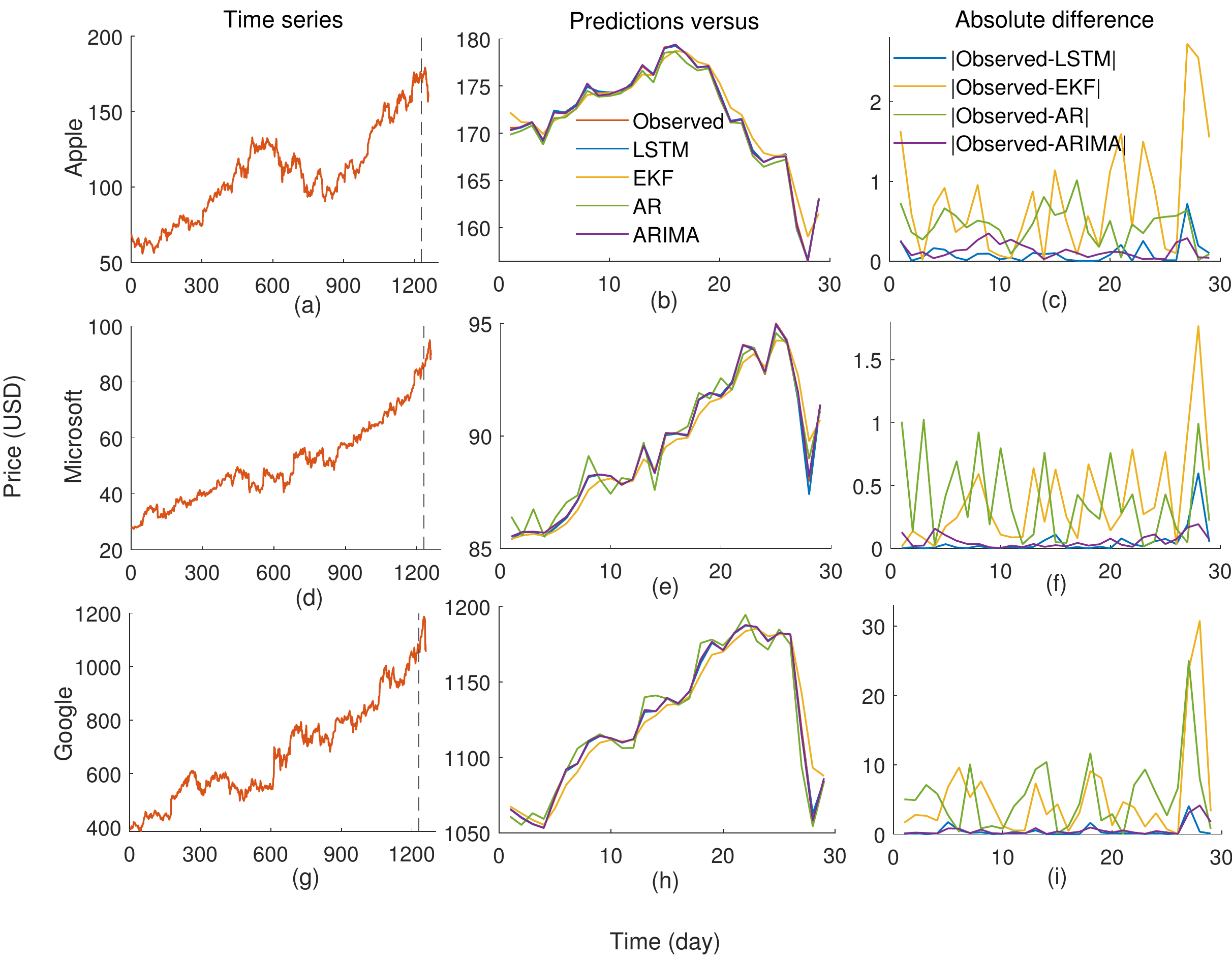}
\caption{Price prediction of three stocks, Apple (first row), Microsoft (second row), and Google (third row), using our real-time many to one LSTM (blue), EKF (yellow), AR (green), and ARIMA (purple). The first, second, and third columns show the entire time series, observed and predicted time series for the last 30 days of the prediction period, and the absolute difference between observed and predicted time series, respectively. We observe that the order of the best to the worst performance is LSTM, ARIMA, AR, and EKF. Note that some of the time series are not visible as they are covered by others.}
 \label{fig:stock}
\end{figure}

Fig.~\ref{fig:stock} shows the price predictions of the three stocks, Apple, Microsoft, and Google, using our real-time many to one LSTM, EKF, AR, and ARIMA. Since some of the predictions closely mimic the observed time series to overlap, we compute the absolute difference between the observations and the predictions, see Figs.~\ref{fig:stock}(c), \ref{fig:stock}(f), and \ref{fig:stock}(i). We observe that all four methods are capable of capturing the pattern of the time series with the order of the best to the worst prediction performance is LSTM, ARIMA, AR, and EKF.  Moreover, while LSTM and ARIMA  perform similarly good, EKF and AR perform similarly weak. 

\begin{figure}[htp]
\centering
\includegraphics[width = 6.5in]{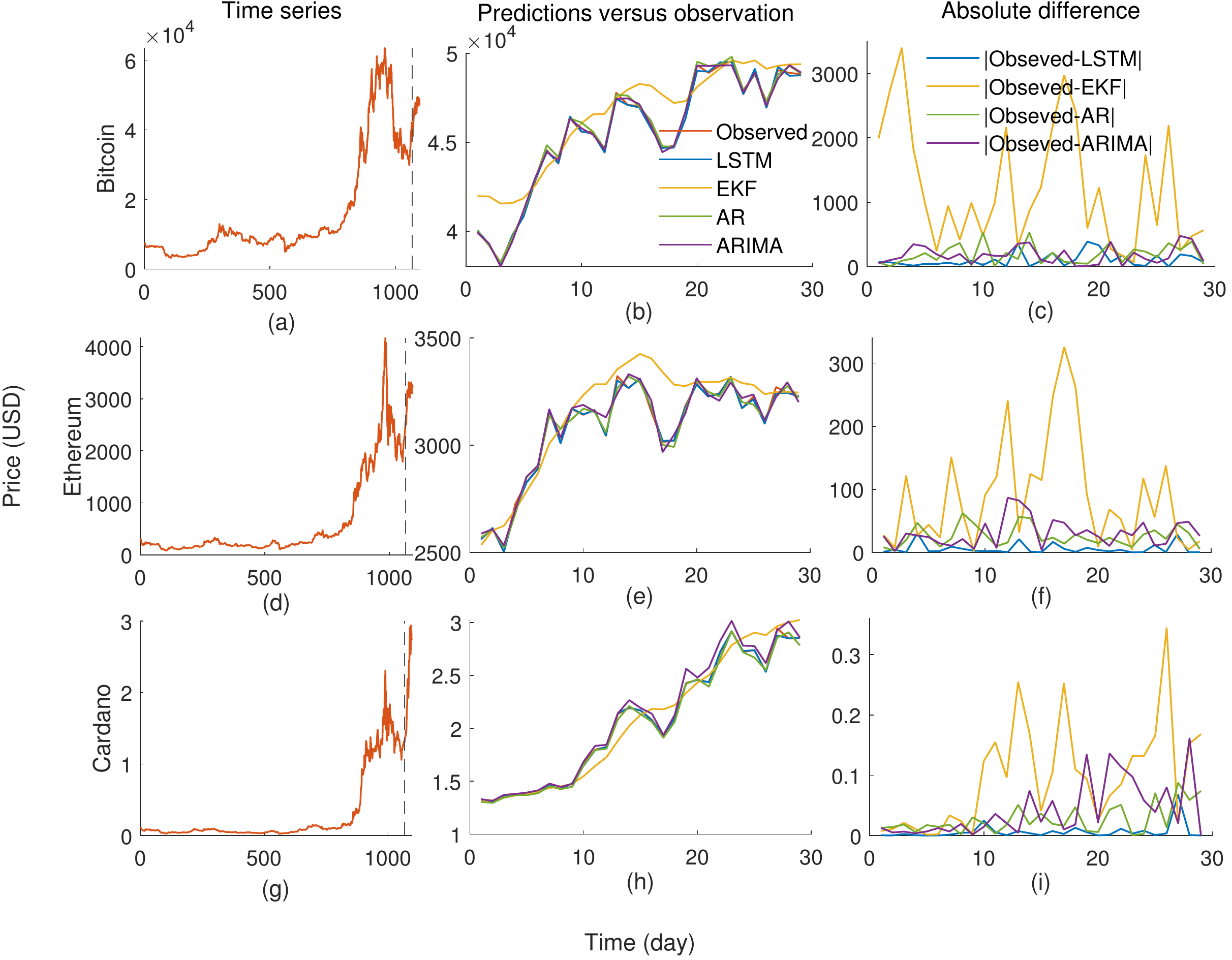}
\caption{Price prediction of three cryptocurrencies, Bitcoin (first row), Ethereum (second row), and Cardano (third row), using our real-time many to one LSTM (blue), EKF (yellow), AR (green), and ARIMA (purple). The first, second, and third columns show the entire time series, observed and predicted time series for the last 30 days of the prediction period, and the absolute difference between observed and predicted time series, respectively. We observe that the order of the best to the worst performance is LSTM, AR, ARIMA, and EKF. Note that some of the time series are not visible as they are covered by others.}
 \label{fig:crypto}
\end{figure}

Fig.~\ref{fig:crypto} shows the price predictions of the three cryptocurrencies, Bitcoin, Ethereum, and Cardano, using LSTM, EKF, AR, and ARIMA. Since some of the predictions closely mimic the observed time series to overlap, we compute the absolute difference between the observations and the predictions, see Figs.~\ref{fig:stock}(c), \ref{fig:stock}(f), and \ref{fig:stock}(i). We observe that LSTM, ARIMA, and AR are capable of capturing the pattern of the time series in contrast to the weak prediction of EKF. The order of the best to the worst prediction performance is LSTM, AR, ARIMA, and EKF.

\begin{figure}[htp]
\centering
\includegraphics[width = 6.5in]{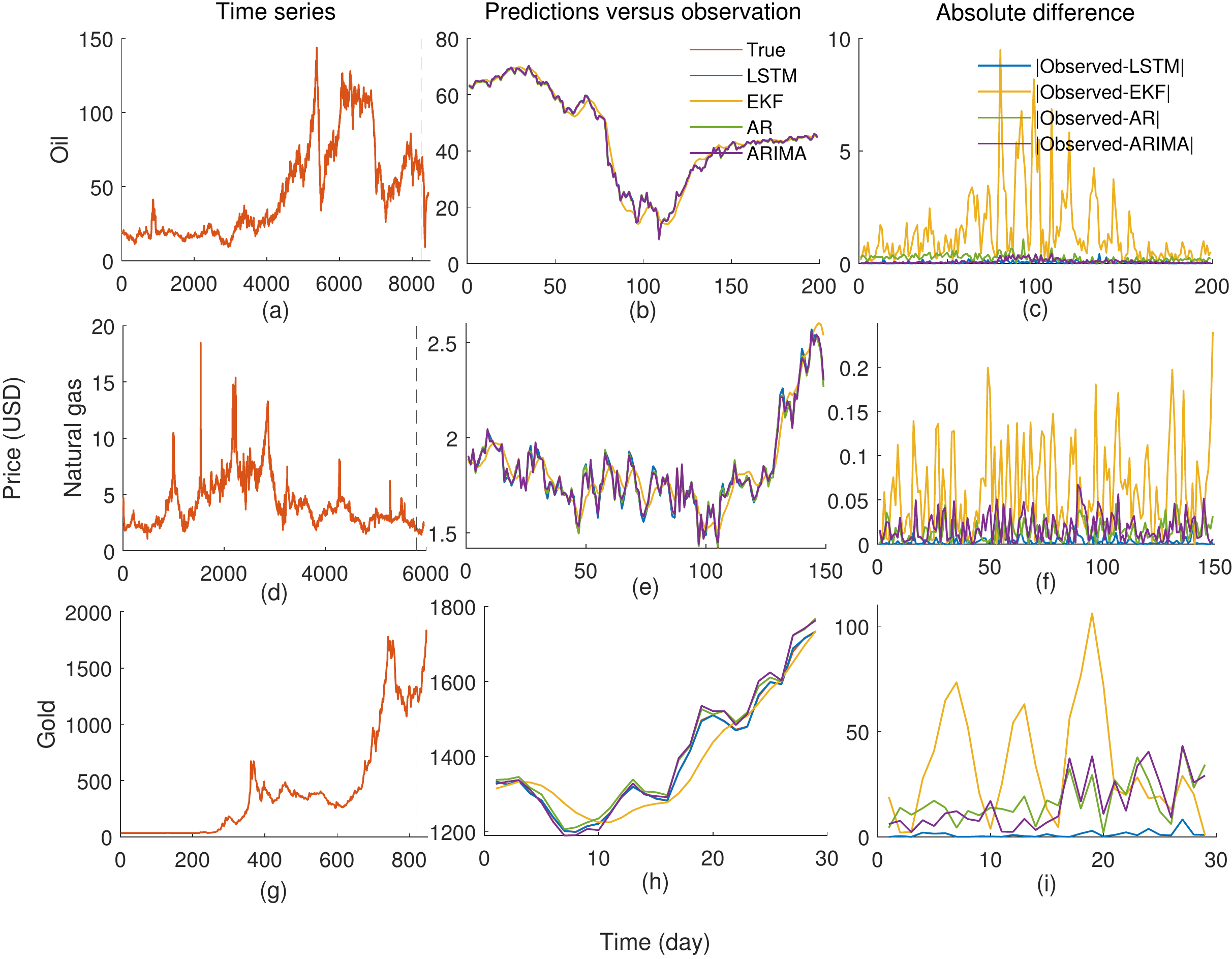}
\caption{Price prediction of three commodities, Oil (first row), Natural gas (second row), and Gold (third row), using our real-time many to one LSTM (blue), EKF (yellow), AR (green), and ARIMA (purple). The first, second, and third columns show the entire time series, observed and predicted time series for the last 30 days of the prediction period, and the absolute difference between observed and predicted time series, respectively. We observe that the order of the best to the worst performance is LSTM, AR, ARIMA, and EKF. Note that some of the time series are not visible as they are covered by others.}
 \label{fig:comm}
\end{figure}

Fig.~\ref{fig:comm} shows the price predictions of the three commodities, Oil, Natural gas, and Gold, using LSTM, EKF, AR, and ARIMA. We compute the absolute difference between the observations and the predictions, see Figs.~\ref{fig:crypto}(c), \ref{fig:crypto}(f), and \ref{fig:crypto}(i) since some of the predictions are similar to observations. We observe that mostly LSTM, ARIMA, and AR are capable of capturing the pattern of the time series. The order of the best to the worst prediction performance is LSTM, Ar, ARIMA, and EKF.

The performance of this real-time many-to-one LSTM is highly influenced by the number of epochs that it is executed. To check this assertion, we compute the prediction performance of the LSTM with respect to different numbers of epochs for Apple, Bitcoin, and Gold. The prediction performance is computed as the mean of the relative absolute difference, i.e., $\mathcal{E}$, between the prediction and the observed time series. Since EKF, AR, and ARIMA are independent of epochs, we represent their $\mathcal{E}$ as a straight line. We observe that the performance of LSTM improves from worst to the best when the number of epochs is increased. 
\begin{figure}[htp]
\includegraphics[width = 6.5in]{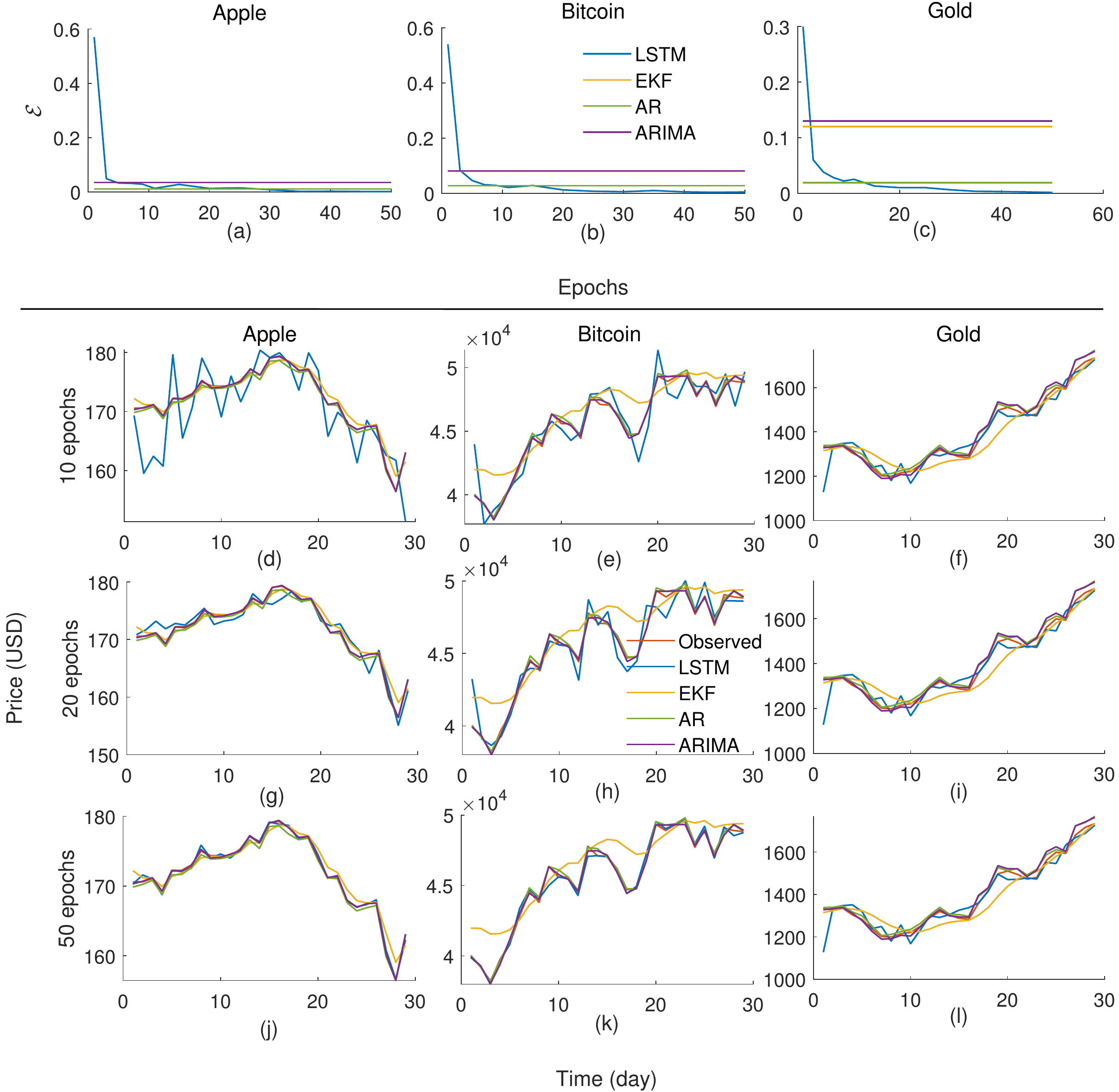}
\caption{Prediction performance of the real-time many-to-one LSTM with respect to the different numbers of epochs. The first row shows the mean of the relative absolute difference, denoted as $\mathcal{E}$, between the prediction and the observed time series for Apple, Bitcoin, and Gold. Note that, EKF, AR, and ARIMA are independent of epochs; however, we represent a straight line for their $\mathcal{E}$ in the first row. The second, the third, and the fourth rows show the prediction and the observed time series of the prices in the prediction periods of Apple, Bitcoin, and Gold for 10 epochs, 20 epochs, and 50 epochs, respectively. We observe that the performance of LSTM improves from worst to the best when the number of epochs is increased. Note that some of the time series and plots are not visible as they are covered by others.}
\label{fig:epoch}
\end{figure}

\section{Discussion} \label{sec:disc}
The classical methods of solving temporal chaotic systems are mostly linear models which assume linear relationships between systems’ previous outputs for stationary time series. Thus, they often do not capture non-linear relationships in the data and cannot cope with certain non-stationary signals. Because financial time series are often nonstationary, nonlinear, and contain noise \citep{bontempi2012machine}, traditional statistical models encounter some limitations in predicting them with high precision. In this paper, we have presented a real-time forecasting technique for financial markets using sequentially trained many-to-one LSTM. We applied this technique for some time series obtained from stock market, cryptocurrency market, and commodity market, then, compared the performance against three state-of-the-art methods, namely, EKF, AR, and ARIMA.

Here, we train a many-to-one LSTM with sequential data sampled using a moving window approach such that the succeeding window is shifted forward by one data instance from the preceding window. Such sequential window training plays an important role in time series predictions since it, 1) helps generate more data from a given limited time series and then thorough training of the ANN; 2) makes the data heterogeneous so that the overfitting issue of the ANN can be reduced; and 3) facilitates the learning of patterns of the data not only for the entire time series but also for short segments of sequential data. Sequential window training maximizes the performance of this LSTM as accelerates LSTM's learning capability as well as it increases LSTM's robustness to new data. 

The performance analysis of this study covers the LSTM applied to nine time series obtained from three financial markets, stocks (Apple, Microsoft, Google), cryptocurrencies (Bitcoin, Ethereum, Cardano), and commodities (gold, crude oil, natural gas). We observed that the LSTM performs exceptionally better than the other three methods for all the nine datasets where the performance of EKF was significantly weak. We have seen in Table~\ref{tab:rmse}, on average, LSTM performs 17 times better than EKF, 7 times better than AR, and 4 times better than ARIMA. The average prediction errors of LSTM are 0.05, 0.22, and 0.14 for stocks, cryptocurrencies, and commodities, respectively. The reason for that is while the prediction on less volatile time series like in the stock market is easy, the prediction on high volatile time series like in the cryptocurrency market is challenging.

In future work, we are planning to extend this sequentially trained many-to-one LSTM to employ as a real-time fault detection technique in industrial production processes. This real-time fault detection scheme will be capable of producing an early alarm to alert a shift in the production process so the quality controlling team can take necessary actions. Moreover, trajectories of collectively moving agents can be represented on a low-dimensional manifold that underlies on a high-dimensional data cloud \cite{SGE,Gajamannage2021,gajamannage2015identifying}. However, some segments of these trajectories are not tracked by multi-object tracking methods due to natural phenomena such as occlusions. Thus, we are planning to utilize our LSTM architecture to make predictions for the fragmented segments of the trajectories. 

We empirically validated that our real-time LSTM outperforms the performance of EKF, AR, and ARIMA. In the future, we are planning to compare the performance of our real-time LSTM with that of the other famous ANN-based methods such as Facebook developed Prophet \citep{taylor2018forecasting}, Amazon developed DeepAR \citep{salinas2020deepar}, Google developed Temporal Fusion Transformer \citep{lim2021temporal}, and Element AI developed N-BEATS \citep{oreshkin2019n}. Prophet was designed for automatic forecasting of univariate time series data. DeepAR is a probabilistic forecasting model based on recurrent neural networks. Temporal Fusion Transformer is a novel attention-based architecture that combines high-performance multi-horizon forecasting with interpretable insights into temporal dynamics. N-BEATS is a custom deep learning algorithm that is based on backward and forward residual links for univariate time series point forecasting. 

We presented both nonlinear and real-time prediction technique for financial time series that is made by a many-to-one LSTM which is sequentially trained with windows of data. The sequential window training approach has significantly improved LSTM's learning ability while dramatically reducing LSTM's over-fitting issues. We empirically justified that our LSTM possesses superior performance even for highly volatile time series such as those in cryptocurrencies and commodities. 

\section*{Acknowledgments}
The authors would like to thank the Google Cloud Platform for granting Research Credit to access its GPU computing resources under project number 397744870419.



\begin{thebibliography}{30}
\expandafter\ifx\csname natexlab\endcsname\relax\def\natexlab#1{#1}\fi
\providecommand{\url}[1]{\texttt{#1}}
\providecommand{\href}[2]{#2}
\providecommand{\path}[1]{#1}
\providecommand{\DOIprefix}{doi:}
\providecommand{\ArXivprefix}{arXiv:}
\providecommand{\URLprefix}{URL: }
\providecommand{\Pubmedprefix}{pmid:}
\providecommand{\doi}[1]{\href{http://dx.doi.org/#1}{\path{#1}}}
\providecommand{\Pubmed}[1]{\href{pmid:#1}{\path{#1}}}
\providecommand{\bibinfo}[2]{#2}
\ifx\xfnm\relax \def\xfnm[#1]{\unskip,\space#1}\fi
\bibitem[{Allen-Zhu et~al.(2019)Allen-Zhu, Li \& Song}]{pascanu2013difficulty}
\bibinfo{author}{Allen-Zhu, Z.}, \bibinfo{author}{Li, Y.}, \&
  \bibinfo{author}{Song, Z.} (\bibinfo{year}{2019}).
\newblock \bibinfo{title}{{On the convergence rate of training recurrent neural
  networks}}.
\newblock In {\it \bibinfo{booktitle}{Advances in Neural Information Processing
  Systems}\/} (pp. \bibinfo{pages}{1310--1318}).
\newblock \bibinfo{organization}{PMLR} volume~\bibinfo{volume}{32}.
\newblock \href{http://arxiv.org/abs/1810.12065}{\tt arXiv:1810.12065}.
\bibitem[{Bahadur et~al.(2019)Bahadur, Paffenroth \& Gajamannage}]{Bahadur2019}
\bibinfo{author}{Bahadur, N.}, \bibinfo{author}{Paffenroth, R.}, \&
  \bibinfo{author}{Gajamannage, K.} (\bibinfo{year}{2019}).
\newblock \bibinfo{title}{{Dimension Estimation of Equity Markets}}.
\newblock In {\it \bibinfo{booktitle}{Proceedings - 2019 IEEE International
  Conference on Big Data, Big Data 2019}\/} (pp. \bibinfo{pages}{5491--5498}).
\newblock \bibinfo{publisher}{Institute of Electrical and Electronics Engineers
  Inc.}
\newblock \DOIprefix\doi{10.1109/BigData47090.2019.9006343}.
\bibitem[{Bontempi et~al.(2013)Bontempi, {Ben Taieb} \& {Le
  Borgne}}]{bontempi2012machine}
\bibinfo{author}{Bontempi, G.}, \bibinfo{author}{{Ben Taieb}, S.}, \&
  \bibinfo{author}{{Le Borgne}, Y.~A.} (\bibinfo{year}{2013}).
\newblock \bibinfo{title}{{Machine learning strategies for time series
  forecasting}}.
\newblock In {\it \bibinfo{booktitle}{Lecture Notes in Business Information
  Processing}\/} (pp. \bibinfo{pages}{62--77}).
\newblock \bibinfo{organization}{Springer} volume \bibinfo{volume}{138 LNBIP}.
\newblock \DOIprefix\doi{10.1007/978-3-642-36318-4_3}.
\bibitem[{Gajamannage et~al.(2015)Gajamannage, Butail, Porfiri \&
  Bollt}]{gajamannage2015identifying}
\bibinfo{author}{Gajamannage, K.}, \bibinfo{author}{Butail, S.},
  \bibinfo{author}{Porfiri, M.}, \& \bibinfo{author}{Bollt, E.~M.}
  (\bibinfo{year}{2015}).
\newblock \bibinfo{title}{{Identifying manifolds underlying group motion in
  Vicsek agents}}.
\newblock {\it \bibinfo{journal}{European Physical Journal: Special Topics}\/},
   {\it \bibinfo{volume}{224}\/}, \bibinfo{pages}{3245--3256}.
  \DOIprefix\doi{10.1140/epjst/e2015-50088-2}.
\bibitem[{Gajamannage \& Paffenroth(2021)}]{Gajamannage2021}
\bibinfo{author}{Gajamannage, K.}, \& \bibinfo{author}{Paffenroth, R.}
  (\bibinfo{year}{2021}).
\newblock \bibinfo{title}{{Bounded manifold completion}}.
\newblock {\it \bibinfo{journal}{Pattern Recognition}\/},  {\it
  \bibinfo{volume}{111}\/}, \bibinfo{pages}{107661}.
  \DOIprefix\doi{https://doi.org/10.1016/j.patcog.2020.107661}.
\bibitem[{Gajamannage et~al.(2019)Gajamannage, Paffenroth \& Bollt}]{SGE}
\bibinfo{author}{Gajamannage, K.}, \bibinfo{author}{Paffenroth, R.}, \&
  \bibinfo{author}{Bollt, E.~M.} (\bibinfo{year}{2019}).
\newblock \bibinfo{title}{{A nonlinear dimensionality reduction framework using
  smooth geodesics}}.
\newblock {\it \bibinfo{journal}{Pattern Recognition}\/},  {\it
  \bibinfo{volume}{87}\/}, \bibinfo{pages}{226--236}.
  \DOIprefix\doi{10.1016/j.patcog.2018.10.020}.
\bibitem[{Gajamannage et~al.(2021)Gajamannage, Park, Paffenroth \&
  Jayasumana}]{gajamannage2022}
\bibinfo{author}{Gajamannage, K.}, \bibinfo{author}{Park, Y.},
  \bibinfo{author}{Paffenroth, R.}, \& \bibinfo{author}{Jayasumana, A.~P.}
  (\bibinfo{year}{2021}).
\newblock \bibinfo{title}{{Reconstruction of Fragmented Trajectories of
  Collective Motion using Hadamard Deep Autoencoders}}.
\newblock {\it \bibinfo{journal}{arXiv preprint arXiv:2110.10428}\/}, .
  \DOIprefix\doi{10.48550/arxiv.2110.10428}.
  \href{http://arxiv.org/abs/2110.10428}{\tt arXiv:2110.10428}.
\bibitem[{Geurts et~al.(1977)Geurts, Box \& Jenkins}]{Geurts1977}
\bibinfo{author}{Geurts, M.}, \bibinfo{author}{Box, G. E.~P.}, \&
  \bibinfo{author}{Jenkins, G.~M.} (\bibinfo{year}{1977}).
\newblock \bibinfo{title}{{Time Series Analysis: Forecasting and Control}}.
\newblock {\it \bibinfo{journal}{Journal of Marketing Research}\/},  {\it
  \bibinfo{volume}{14}\/}, \bibinfo{pages}{269}.
  \DOIprefix\doi{10.2307/3150485}.
\bibitem[{Gruslys et~al.(2016)Gruslys, Munos, Danihelka, Lanctot \&
  Graves}]{gruslys2016memory}
\bibinfo{author}{Gruslys, A.}, \bibinfo{author}{Munos, R.},
  \bibinfo{author}{Danihelka, I.}, \bibinfo{author}{Lanctot, M.}, \&
  \bibinfo{author}{Graves, A.} (\bibinfo{year}{2016}).
\newblock \bibinfo{title}{{Memory-efficient backpropagation through time}}.
\newblock {\it \bibinfo{journal}{Advances in Neural Information Processing
  Systems}\/},  {\it \bibinfo{volume}{29}\/}, \bibinfo{pages}{4132--4140}.
  \href{http://arxiv.org/abs/1606.03401}{\tt arXiv:1606.03401}.
\bibitem[{Hochreiter \& Schmidhuber(1997)}]{hochreiter1997long}
\bibinfo{author}{Hochreiter, S.}, \& \bibinfo{author}{Schmidhuber, J.}
  (\bibinfo{year}{1997}).
\newblock \bibinfo{title}{{Long short-term memory}}.
\newblock {\it \bibinfo{journal}{Neural computation}\/},  {\it
  \bibinfo{volume}{9}\/}, \bibinfo{pages}{1735--1780}.
  \DOIprefix\doi{10.1162/neco.1997.9.8.1735}.
\bibitem[{Kim et~al.(2017)Kim, {El Khamy} \& Lee}]{kim2017residual}
\bibinfo{author}{Kim, J.}, \bibinfo{author}{{El Khamy}, M.}, \&
  \bibinfo{author}{Lee, J.} (\bibinfo{year}{2017}).
\newblock \bibinfo{title}{{Residual LSTM: Design of a deep recurrent
  architecture for distant speech recognition}}.
\newblock {\it \bibinfo{journal}{Proceedings of the Annual Conference of the
  International Speech Communication Association}\/},  {\it
  \bibinfo{volume}{2017-Augus}\/}, \bibinfo{pages}{1591--1595}.
  \DOIprefix\doi{10.21437/Interspeech.2017-477}.
\bibitem[{Kingma \& Ba(2015)}]{kingma2014adam}
\bibinfo{author}{Kingma, D.~P.}, \& \bibinfo{author}{Ba, J.~L.}
  (\bibinfo{year}{2015}).
\newblock \bibinfo{title}{{Adam: A method for stochastic optimization}}.
\newblock {\it \bibinfo{journal}{3rd International Conference on Learning
  Representations, ICLR 2015 - Conference Track Proceedings}\/}, .
  \DOIprefix\doi{10.48550/arXiv.1412.6980}.
  \href{http://arxiv.org/abs/1412.6980}{\tt arXiv:1412.6980}.
\bibitem[{Le \& Zuidema(2016)}]{le2016quantifying}
\bibinfo{author}{Le, P.}, \& \bibinfo{author}{Zuidema, W.}
  (\bibinfo{year}{2016}).
\newblock \bibinfo{title}{{Quantifying the Vanishing Gradient and Long Distance
  Dependency Problem in Recursive Neural Networks and Recursive LSTMs}}.
\newblock {\it \bibinfo{journal}{arXiv preprint arXiv:1603.00423}\/},  (pp.
  \bibinfo{pages}{87--93}). \DOIprefix\doi{10.18653/v1/w16-1610}.
  \href{http://arxiv.org/abs/1603.00423}{\tt arXiv:1603.00423}.
\bibitem[{Lim et~al.(2021)Lim, Arık, Loeff \& Pfister}]{lim2021temporal}
\bibinfo{author}{Lim, B.}, \bibinfo{author}{Arık, S.}, \bibinfo{author}{Loeff,
  N.}, \& \bibinfo{author}{Pfister, T.} (\bibinfo{year}{2021}).
\newblock \bibinfo{title}{{Temporal Fusion Transformers for interpretable
  multi-horizon time series forecasting}}.
\newblock {\it \bibinfo{journal}{International Journal of Forecasting}\/},
  {\it \bibinfo{volume}{37}\/}, \bibinfo{pages}{1748--1764}.
  \DOIprefix\doi{10.1016/j.ijforecast.2021.03.012}.
  \href{http://arxiv.org/abs/1912.09363}{\tt arXiv:1912.09363}.
\bibitem[{Ma \& Principe(2018)}]{ma2018comparison}
\bibinfo{author}{Ma, Y.}, \& \bibinfo{author}{Principe, J.}
  (\bibinfo{year}{2018}).
\newblock \bibinfo{title}{{Comparison of Static Neural Network with External
  Memory and RNNs for Deterministic Context Free Language Learning}}.
\newblock In {\it \bibinfo{booktitle}{Proceedings of the International Joint
  Conference on Neural Networks}\/} (pp. \bibinfo{pages}{1--7}).
\newblock \bibinfo{organization}{IEEE} volume \bibinfo{volume}{2018-July}.
\newblock \DOIprefix\doi{10.1109/IJCNN.2018.8489240}.
\bibitem[{Manneschi \& Vasilaki(2020)}]{manneschi2020alternative}
\bibinfo{author}{Manneschi, L.}, \& \bibinfo{author}{Vasilaki, E.}
  (\bibinfo{year}{2020}).
\newblock \bibinfo{title}{{An alternative to backpropagation through time}}.
\newblock {\it \bibinfo{journal}{Nature Machine Intelligence}\/},  {\it
  \bibinfo{volume}{2}\/}, \bibinfo{pages}{155--156}.
  \DOIprefix\doi{10.1038/s42256-020-0162-9}.
\bibitem[{Newbold(1983)}]{newbold1983arima}
\bibinfo{author}{Newbold, P.} (\bibinfo{year}{1983}).
\newblock \bibinfo{title}{{ARIMA model building and the time series analysis
  approach to forecasting}}.
\newblock {\it \bibinfo{journal}{Journal of Forecasting}\/},  {\it
  \bibinfo{volume}{2}\/}, \bibinfo{pages}{23--35}.
  \DOIprefix\doi{10.1002/for.3980020104}.
\bibitem[{Oreshkin et~al.(2019)Oreshkin, Carpov, Chapados \&
  Bengio}]{oreshkin2019n}
\bibinfo{author}{Oreshkin, B.~N.}, \bibinfo{author}{Carpov, D.},
  \bibinfo{author}{Chapados, N.}, \& \bibinfo{author}{Bengio, Y.}
  (\bibinfo{year}{2019}).
\newblock \bibinfo{title}{{N-BEATS: Neural basis expansion analysis for
  interpretable time series forecasting}}.
\newblock {\it \bibinfo{journal}{arXiv preprint arXiv:1905.10437}\/}, .
  \URLprefix \url{http://arxiv.org/abs/1905.10437}.
  \href{http://arxiv.org/abs/1905.10437}{\tt arXiv:1905.10437}.
\bibitem[{Park et~al.(2022)Park, Gajamannage, Jayathilake \&
  Bollt}]{Yonggi2022}
\bibinfo{author}{Park, Y.}, \bibinfo{author}{Gajamannage, K.},
  \bibinfo{author}{Jayathilake, D.~I.}, \& \bibinfo{author}{Bollt, E.~M.}
  (\bibinfo{year}{2022}).
\newblock \bibinfo{title}{{Recurrent Neural Networks for Dynamical Systems :
  Applications to Ordinary Differential Equations , Collective Motion , and
  Hydrological Modeling}}, .
\newblock (pp. \bibinfo{pages}{1--15}).
  \DOIprefix\doi{10.48550/arxiv.2202.07022}.
\bibitem[{Qiu et~al.(2020)Qiu, Wang \& Zhou}]{qiu2020forecasting}
\bibinfo{author}{Qiu, J.}, \bibinfo{author}{Wang, B.}, \&
  \bibinfo{author}{Zhou, C.} (\bibinfo{year}{2020}).
\newblock \bibinfo{title}{{Forecasting stock prices with long-short term memory
  neural network based on attention mechanism}}.
\newblock {\it \bibinfo{journal}{PLoS ONE}\/},  {\it \bibinfo{volume}{15}\/},
  \bibinfo{pages}{e0227222}. \DOIprefix\doi{10.1371/journal.pone.0227222}.
\bibitem[{Salinas et~al.(2020)Salinas, Flunkert, Gasthaus \&
  Januschowski}]{salinas2020deepar}
\bibinfo{author}{Salinas, D.}, \bibinfo{author}{Flunkert, V.},
  \bibinfo{author}{Gasthaus, J.}, \& \bibinfo{author}{Januschowski, T.}
  (\bibinfo{year}{2020}).
\newblock \bibinfo{title}{{DeepAR: Probabilistic forecasting with
  autoregressive recurrent networks}}.
\newblock {\it \bibinfo{journal}{International Journal of Forecasting}\/},
  {\it \bibinfo{volume}{36}\/}, \bibinfo{pages}{1181--1191}.
  \DOIprefix\doi{10.1016/j.ijforecast.2019.07.001}.
  \href{http://arxiv.org/abs/1704.04110}{\tt arXiv:1704.04110}.
\bibitem[{Shih et~al.(2018)Shih, Yan, Liu \& Chen}]{shih2017investigating}
\bibinfo{author}{Shih, C.~H.}, \bibinfo{author}{Yan, B.~C.},
  \bibinfo{author}{Liu, S.~H.}, \& \bibinfo{author}{Chen, B.}
  (\bibinfo{year}{2018}).
\newblock \bibinfo{title}{{Investigating Siamese LSTM networks for text
  categorization}}.
\newblock In {\it \bibinfo{booktitle}{Proceedings - 9th Asia-Pacific Signal and
  Information Processing Association Annual Summit and Conference, APSIPA ASC
  2017}\/} (pp. \bibinfo{pages}{641--646}).
\newblock \bibinfo{organization}{IEEE} volume \bibinfo{volume}{2018-Febru}.
\newblock \DOIprefix\doi{10.1109/APSIPA.2017.8282104}.
\bibitem[{Shumway \& Stoffer(2017)}]{arima}
\bibinfo{author}{Shumway, R.~H.}, \& \bibinfo{author}{Stoffer, D.~S.}
  (\bibinfo{year}{2017}).
\newblock \bibinfo{title}{{ARIMA Models}}.
\newblock (pp. \bibinfo{pages}{75--163}).
\newblock \bibinfo{publisher}{Springer, Cham}.
\newblock \DOIprefix\doi{10.1007/978-3-319-52452-8_3}.
\bibitem[{Simistira et~al.(2015)Simistira, Ul-Hassan, Papavassiliou, Gatos,
  Katsouros \& Liwicki}]{simistira2015recognition}
\bibinfo{author}{Simistira, F.}, \bibinfo{author}{Ul-Hassan, A.},
  \bibinfo{author}{Papavassiliou, V.}, \bibinfo{author}{Gatos, B.},
  \bibinfo{author}{Katsouros, V.}, \& \bibinfo{author}{Liwicki, M.}
  (\bibinfo{year}{2015}).
\newblock \bibinfo{title}{{Recognition of historical Greek polytonic scripts
  using LSTM networks}}.
\newblock In {\it \bibinfo{booktitle}{Proceedings of the International
  Conference on Document Analysis and Recognition, ICDAR}\/} (pp.
  \bibinfo{pages}{766--770}).
\newblock \bibinfo{organization}{IEEE} volume \bibinfo{volume}{2015-Novem}.
\newblock \DOIprefix\doi{10.1109/ICDAR.2015.7333865}.
\bibitem[{Squarepants(2022)}]{nakamoto2008bitcoin}
\bibinfo{author}{Squarepants, S.} (\bibinfo{year}{2022}).
\newblock \bibinfo{title}{{Bitcoin: A Peer-to-Peer Electronic Cash System}}.
\newblock {\it \bibinfo{journal}{SSRN Electronic Journal}\/},  (p.
  \bibinfo{pages}{21260}). \DOIprefix\doi{10.2139/ssrn.3977007}.
\bibitem[{Taylor \& Letham(2018)}]{taylor2018forecasting}
\bibinfo{author}{Taylor, S.~J.}, \& \bibinfo{author}{Letham, B.}
  (\bibinfo{year}{2018}).
\newblock \bibinfo{title}{{Forecasting at Scale}}.
\newblock {\it \bibinfo{journal}{American Statistician}\/},  {\it
  \bibinfo{volume}{72}\/}, \bibinfo{pages}{37--45}.
  \DOIprefix\doi{10.1080/00031305.2017.1380080}.
\bibitem[{Tian et~al.(2017)Tian, Zhang, Ma, He, Wei, Wu, Situ, Li \&
  Zhang}]{tian2017deep}
\bibinfo{author}{Tian, X.}, \bibinfo{author}{Zhang, J.}, \bibinfo{author}{Ma,
  Z.}, \bibinfo{author}{He, Y.}, \bibinfo{author}{Wei, J.},
  \bibinfo{author}{Wu, P.}, \bibinfo{author}{Situ, W.}, \bibinfo{author}{Li,
  S.}, \& \bibinfo{author}{Zhang, Y.} (\bibinfo{year}{2017}).
\newblock \bibinfo{title}{{Deep LSTM for large vocabulary continuous speech
  recognition}}, .
\newblock \DOIprefix\doi{10.48550/arXiv.1703.07090}.
  \href{http://arxiv.org/abs/1703.07090}{\tt arXiv:1703.07090}.
\bibitem[{Valade et~al.(2017)Valade, Acco, Grabolosa \&
  Fourniols}]{valade2017study}
\bibinfo{author}{Valade, A.}, \bibinfo{author}{Acco, P.},
  \bibinfo{author}{Grabolosa, P.}, \& \bibinfo{author}{Fourniols, J.~Y.}
  (\bibinfo{year}{2017}).
\newblock \bibinfo{title}{{A study about kalman filters applied to embedded
  sensors}}.
\newblock {\it \bibinfo{journal}{Sensors (Switzerland)}\/},  {\it
  \bibinfo{volume}{17}\/}, \bibinfo{pages}{2810}.
  \DOIprefix\doi{10.3390/s17122810}.
\bibitem[{Werbos(1990)}]{werbos1990backpropagation}
\bibinfo{author}{Werbos, P.~J.} (\bibinfo{year}{1990}).
\newblock \bibinfo{title}{{Backpropagation Through Time: What It Does and How
  to Do It}}.
\newblock {\it \bibinfo{journal}{Proceedings of the IEEE}\/},  {\it
  \bibinfo{volume}{78}\/}, \bibinfo{pages}{1550--1560}.
  \DOIprefix\doi{10.1109/5.58337}.
\bibitem[{Zhao et~al.(2018)Zhao, Ge, Zhou, Sun, Zheng, Wang, Huang \&
  Cheng}]{zhao2018new}
\bibinfo{author}{Zhao, Y.}, \bibinfo{author}{Ge, L.}, \bibinfo{author}{Zhou,
  Y.}, \bibinfo{author}{Sun, Z.}, \bibinfo{author}{Zheng, E.},
  \bibinfo{author}{Wang, X.}, \bibinfo{author}{Huang, Y.}, \&
  \bibinfo{author}{Cheng, H.} (\bibinfo{year}{2018}).
\newblock \bibinfo{title}{{A new Seasonal Difference Space-Time Autoregressive
  Integrated Moving Average (SD-STARIMA) model and spatiotemporal trend
  prediction analysis for Hemorrhagic Fever with Renal Syndrome (HFRS)}}.
\newblock {\it \bibinfo{journal}{PLoS ONE}\/},  {\it \bibinfo{volume}{13}\/},
  \bibinfo{pages}{e0207518}. \DOIprefix\doi{10.1371/journal.pone.0207518}.

\end{thebibliography}

\end{document}